\documentclass[twoside]{article}

\usepackage[accepted]{aistats2019}
\usepackage[round]{natbib}

\usepackage{amssymb}
\usepackage{bm}
\usepackage[pdftex]{graphicx}
\usepackage{subcaption}
\usepackage{booktabs}       
\usepackage{url}            

\renewcommand{\vec}[1]{\mathbf{#1}}
\newcommand{\boldpar}[1]{\textbf{#1. }}

\begin{document}

%

%

\twocolumn[

\aistatstitle{Inverting Supervised Representations with Autoregressive Neural Density Models}
\aistatsauthor{ Charlie Nash \And Nate Kushman \And  Christopher K.I. Williams }
\aistatsaddress{ 
	School of Informatics \\ 
	University of Edinburgh, UK \\
	\texttt{charlie.nash@ed.ac.uk}
	\And
	Microsoft Research \\
	Cambridge, UK \\
	\texttt{nkushman@microsoft.com}
	\And
	School of Informatics \\
	University of Edinburgh, UK \\
	Alan Turing Institute, London, UK \\
	\texttt{ckiw@inf.ed.ac.uk}
} 
]
\begin{abstract}
 We present a method for feature interpretation that makes use of recent advances in autoregressive density estimation models to invert model representations. We train generative \emph{inversion models} to express a distribution over input features conditioned on intermediate model representations. Insights into the invariances learned by supervised models can be gained by viewing samples from these inversion models. In addition, we can use these inversion models to estimate the mutual information between a model's inputs and its intermediate representations, thus quantifying the amount of information preserved by the network at different stages.  Using this method we examine the types of information preserved at different layers of convolutional neural networks, and explore the invariances induced by different architectural choices. Finally we show that the mutual information between inputs and network layers initially increases and then decreases over the course of training, supporting recent work by Shwartz-Ziv and Tishby \citeyearpar{Shwartz-ZivT17} on the information bottleneck theory of deep learning.
\end{abstract}

\section{Introduction}\label{sec:intro}The representations learned in supervised models are task specific; they discard irrelevant input information and preserve features that are useful for characterizing their targets. This is the conventional wisdom taken for granted by many in the machine learning community. However the precise nature of what information is preserved across different layers of a neural network is generally unknown. A better understanding of this is desirable both for the interpretability of a particular network, and for the insights that can be gained for neural architecture design. 

We expect supervised models to be invariant to certain transformations of the input data. For instance an effective image classifier should be invariant to translations of objects in the image, and such behaviour is encouraged through architectural choices like convolutions and pooling. As such we anticipate that the mapping from inputs to intermediate representations discards information, and that perfect recovery of the inputs is not possible. Recent work by Schwartz-Ziv and Tishby \citeyearpar{Shwartz-ZivT17} argues that compression of input data in network representations is a central reason for the success of deep models, particularly with respect to generalization performance. Despite this, attempts such as \citep{DosovitskiyB16a} have been made to invert representations using reconstructions that are optimized to minimize pixel losses such as mean squared error. This leads to blurry reconstructions from higher-level representations. Perceptual losses using features from a pretrained convolutional network or adversarial discriminator networks significantly improve the visual quality of results \citep{DosovitskiyB16,JohnsonAF16}, but fail to characterize the variability present in the inverse mapping. We propose instead to express a distribution over the inputs conditioned on a network representation. By sampling from this conditional distribution we can visualize the types of inputs that map to a given representation. 

In recent years there have been significant advances in neural generative models of high-dimensional data \citep{Goodfellow14,Kingma14,RezendeMW14}. \emph{Autoregressive} density models decompose a joint distribution into products of conditionals. By leveraging domain-specific structure, impressive results have been achieved with neural autoregressive models of images \citep{OordKK16,OordKEKVG16} and audio \citep{OordDZSVGKSK16}.  Unlike alternative generative models such as variational autoencoders \citep{Kingma14,RezendeMW14} or generative adversarial networks \citep{Goodfellow14}, autoregressive models yield an exact density. We show in Section \ref{sec:bounding_the_mutual_information} that this density can be used to estimate a lower bound on the mutual information between inputs and model representations, which is a useful metric for the analysis of neural networks. In contrast to other methods for mutual information estimation this estimate is scalable to high-dimensionality inputs and network features with complex dependencies. In addition autoregressive models are straightforward to train in comparison to other generative models, with a single optimization objective and none of the instability associated with adversarial training. Autoregressive density models are therefore a strong choice for our desired goal of representation inversion. 

In this work we present a method for the inversion of supervised representations that uses flexible autoregressive neural density models to express a distribution over inputs given an intermediate representation. We show how such models can be used to help understand how much and what kind of information is preserved at different hidden layers on a range of image datasets (Sec. \ref{sec:inverting_the_layers}). We use inversion models to visualise the invariances learned by classifiers with different architectures, and demonstrate advantages in interpretability compared to point-estimate approaches (Sec. \ref{sec:case_study}). Finally we demonstrate that the mutual information between inputs and intermediate representations initially increases before decreasing over the course of training, reproducing the results of Schwartz-Ziv and Tishby \citeyearpar{Shwartz-ZivT17} in the context of ReLU-convolutional networks (Sec. \ref{sec:compression_dynamics}).
\section{Related work}\label{sec:related}Our approach is related to many previous works on inverting neural networks. Although our approach has similar goals to optimization-based approaches to network inversion \citep{Linden89, LeeK94, LuKN99, MahendranV15} it is most closely related to methods which make use of another neural network that is trained to invert the hidden states \citep{DosovitskiyB16a,DosovitskiyB16,JohnsonAF16,HuangLPHB17}. In another related work Zeiler and Fergus \citeyearpar{zeiler14} invert individual features by explicitly reversing the filtering, pooling and rectification operations in convolutional networks. Our work is distinct in its use of an autoregressive model to express a distribution over a network's inputs.

Dosovitskiy and Brox \citeyearpar{DosovitskiyB16a} train an up-sampling convolutional network to map from a representation layer $\vec{h}$ to inputs $\hat{\vec{x}} = \vec{f}(\vec{h})$, and optimize the mean squared error with respect to the true inputs $\mathbb{E}_{\vec{x}} || \vec{x} - \vec{f}(\vec{h}) ||^2$. With this method reconstructions become increasingly blurry as the amount of information preserved by the network about the inputs decreases with successive layers. The level of blurriness is quite useful as an indication of the amount of information preserved by the network, however it is a coarse measure that doesn't demonstrate the variability of inputs consistent with a given representation. In addition, this approach is only appropriate for data in continuous spaces where mean squared error is a meaningful metric. Our method can be applied in any setting in which a distribution over inputs can be parameterized, such as language processing.

Johnson et al.~\citeyearpar{JohnsonAF16} augment a pixel loss with perceptual losses that make use of the feature space of a pre-trained classifier to invert VGG features. These additional constraints result in outputs that are visually appealing in comparison to simple pixel losses. Dosovitskiy and Brox \citeyearpar{DosovitskiyB16} extend this approach with an adversarial loss that encourages reconstructed inputs to additionally "fool" a GAN discriminator. Although these approaches produce high quality outputs they are limited in that they produce a single image reconstruction for a given representation, rather than providing a distribution over plausible inputs consistent with the representation. 

Our method is also similar to stacked generative adversarial networks \citep{HuangLPHB17}, in which a series of GANs are used to map from higher to lower level representations of a pre-trained classifier. This model was presented primarily as a way to make use of supervised representations in order to improve sample quality. In order to avoid degenerate samples, an entropy loss was incorporated that encourages the auxiliary noise to be recoverable from samples. This loss provides a lower bound on the entropy of the conditional distributions, and results in diverse samples. However entropy maximization is not equivalent to maximizing the likelihood, and may result in poorly calibrated distributions.

Van den Oord et al.~\citeyearpar{OordKEKVG16} use a conditional PixelCNN to generate images conditioned on portrait embeddings obtained from the top layer of a face-detection CNN trained with triplet loss on Flickr images. This is equivalent to our method, although in that case the emphasis was on portrait generation rather than analysis of the learned representations.
\section{Background}\label{sec:background}\subsection{Autoregressive neural density models}
Neural density models use neural networks to describe parametric distributions $p_{\theta}$ over random variables $\vec{x}$. Autoregressive models decompose the joint distribution into a series of $D$ conditionals $p_{\theta}(\vec{x}) = \prod_i p_{\theta}(x_i | \vec{x}_{1:i-1})$ where the parameters for the $i$'th conditional distribution are the outputs of a network $\bm{\theta}_i = \vec{f}(\vec{x}_{1:i-1})$ that takes the preceding variables as input. Density models are typically trained to maximize the likelihood with respect to samples from the true data distribution. Various neural density models have been proposed; from general purpose models \citep{UriaML13,GermainGML15,PapamakariosPM17} to domain specific models for images \citep{OordKK16,OordKEKVG16}, text \citep{sundermeyer2012}, and audio \citep{OordDZSVGKSK16}. Many neural density models make use of architectures that parallelize the computation of the $D$ conditional distributions through a single pass of a network. This enables efficient computation as well as parameter sharing across conditional distributions. In order to ensure that each conditional only depends on the preceding variables, architectural tools such as causal convolutions and masking are used. 

Conditional density models aim to model a conditional distribution $p(\vec{x} | \vec{h})$ of data variables $\vec{x}$ given context $\vec{h}$. Typical examples include models of images conditioned on object classes, or speech models conditioned on speaker identity. Usually each conditional is allowed to depend on the context such that $p_{\theta}(\vec{x} | \vec{h}) = \prod_i p_{\theta}(x_i | \vec{x}_{1:i-1}, \vec{h})$. We make use of conditional density models in order to model the distribution over input data conditioned on supervised representations.
\subsection{PixelCNN}
The PixelCNN \citep{OordKK16} is an autoregressive neural density model for images that uses a convolutional neural network to parameterize conditional distributions for each sub-pixel in an image. Pixel values are sampled one at a time: from left to right and from top to bottom. Causality in the conditional distributions is maintained using masked convolutions that only allow connections from previously observed pixels. The PixelCNN and its variants \citep{OordKK16, OordKEKVG16, SalimansKCK17} are powerful models of images, and currently are the state of the art with respect to log-likelihood scores on natural images. In our experiments we make use of the PixelCNN++ \citep{SalimansKCK17}, which incorporates a number of changes to the original model including the use of an alternative mixture-based pixel likelihood, downsampling to increase receptive field sizes and short-cut connections. Conditioning information is incorporated by regressing a context vector to biases which are added to intermediate feature maps. For full details see Salimans et al., \citeyearpar{SalimansKCK17} and the implementation at \texttt{https://github.com/openai/pixel-cnn}.

\subsection{Mutual information estimation in neural networks}\label{subsec:mi_estimation}
A quantity of particular interest in the analysis of network representations is the mutual information $I(\vec{x}; \vec{h})$ between inputs $\vec{x}$ and a model representation $\vec{h}$:
\begin{align}
I(\vec{x}; \vec{h}) = H(\vec{x}) - H(\vec{x} | \vec{h}) = H(\vec{h}) - H(\vec{h} | \vec{x}),
\end{align}
where $H(\vec{x})$ denotes the entropy of $\vec{x}$. The mutual information represents the reduction in uncertainty about $\vec{x}$ that we obtain if we know $\vec{h}$, and can be thought of in this context as the amount of information preserved in the transformation $\vec{x} \to \vec{h}$. In general we are unable to obtain this quantity as we don't have access to the true distributions $p(\vec{x})$, $p(\vec{h})$, $p(\vec{x} | \vec{h})$ or $p(\vec{h} | \vec{x})$. However, the mutual information can be approximated in various ways. In this section we focus on mutual information estimation via density estimation. For more details on alternative approaches see Appendix B. It is important to first consider the implications of working with discrete vs continuous probability distributions.

\boldpar{Discrete vs continuous entropy} For deterministic functions of continuous inputs the conditional distribution $p(\vec{h} | \vec{x})$ is degenerate and so the differential entropy $H(\vec{h} | \vec{x})$ is negative infinity. For finite $H(\vec{h})$ this implies that the mutual information is infinite. This poses a problem for the analysis of neural networks with continuous inputs, as network representations are typically a deterministic function of the inputs. In this work we deal exclusively with discrete image inputs, and can avoid this issue by using discrete entropy for which $H(\vec{h} | \vec{x})$ is zero rather than infinite. However, for models with continuous inputs care must be taken to either add noise or to discretize the continuous space. For a more detailed discussion of related issues see Saxe et al.~\citeyearpar{michael2018}.

\boldpar{Density estimation} For networks operating with continuous input spaces one method for mutual information estimation is to add noise to network activations $\vec{h}^{\epsilon} = \vec{h} + \bm{\epsilon}$ and use a parametric or non-parametric model $p^*$ to estimate $p(\vec{h}^{\epsilon})$.  As $\vec{h}$ is a deterministic function of $\vec{x}$ the conditional entropy $H(\vec{h} | \vec{x})$ is simply equal to the entropy of the Gaussian noise $H(\bm{\epsilon})$. The approximate model can then be used to estimate the cross entropy $H(p(\vec{h}^{\epsilon}), p^*(\vec{h}^{\epsilon}))$. An upper bound on the mutual information can then be established as follows:
\begin{align}
I(\vec{x}; \vec{h}^\epsilon) = H(\vec{h^{\epsilon}}) - H(\bm{\epsilon}) \leq H(p(\vec{h}^{\epsilon}), p^*(\vec{h}^{\epsilon})) - c,
\end{align}
where $c = H(\bm{\epsilon})$, and we use the fact that $H(p(\vec{h}^{\epsilon}), p^*(\vec{h}^{\epsilon})) \geq H(\vec{h}^{\epsilon})$. Kolchinksy \& Tracey \citeyearpar{KolchinskyT17}; Kolchinksy et al.~\citeyearpar{KolchinskyTW17} use a kernel density estimate (KDE) of $p(\vec{h}^{\epsilon})$ to obtain a bound on the mutual information as above. We note that the performance of kernel density estimates deteriorate significantly in higher dimensions. Theis et al.~\citeyearpar{Theis2016a} show that even for a very large number of samples, kernel density methods greatly underestimate the true log-likelihood of simple models trained on $6 \times 6$ image patches. As such these methods are not appropriate for large networks. A parametric estimate $p_\theta(\vec{h}^\epsilon)$ using e.g. neural autoregressive density models would potentially scale better to higher dimensions, although to our knowledge this hasn't been explored in previous work.

\section{Inverting Supervised Representations}\label{sec:inverting}Previous approaches to representation inversion \citep{DosovitskiyB16} optimize a parameterized inversion function $\vec{f}_{\theta}$ with respect to the mean squared error of an image and its reconstruction:
\begin{align} \label{eq:mse_loss}
\mathcal{L}_{\theta}^{\text{MSE}} = \mathbb{E}_{p(\vec{x}, \vec{h})} || \vec{x} - \vec{f}_{\theta}(\vec{h}) ||^2.
\end{align}
This use of a point estimate in order to invert an information-lossy transformation results in blurry reconstructions and does not provide information about the variability inherent in the inverse mapping. Our method instead minimizes the negative log-likelihood of a parameterized \emph{inversion} model:
\begin{align}
\mathcal{L}_{\theta}^{\text{NLL}} = -\mathbb{E}_{p(\vec{x}, \vec{h})} [\text{log} \ p_{\theta}(\vec{x} | \vec{h})].
\end{align}
This enables us to query whether a given input is a good match for a particular representation. In addition we can sample our trained model to get a sense of the degree of constraint present in $\vec{h}$. Optimization of Equation \ref{eq:mse_loss} is equivalent to our proposed maximum likelihood criterion if the conditional distribution $p_{\theta}$ is a Gaussian with spherical covariance. As such our method simply extends Equation \ref{eq:mse_loss} by using a more flexible class of conditional probability models. As we have chosen to focus on supervised models of images, we use a conditional variant of the PixelCNN++ as our inversion model. Although we do not explore it here, we note that our method is not tied to any particular density model, and that equivalent domain-appropriate models could be used for e.g. text classification or speech recognition.

\subsection{Bounding the mutual information}\label{sec:bounding_the_mutual_information}
Inversion models can be used to compute an upper bound on $H(\vec{x} | \vec{h})$ using the conditional cross entropy between $p(\vec{x}|\vec{h})$ and the inversion model distribution $p_\theta(\vec{x}|\vec{h})$:
\begin{align}
H(p(\vec{x} & | \vec{h}), p_{\theta}(\vec{x} | \vec{h})) = - \mathbb{E}_{p(\vec{x}, \vec{h})} \left [\text{log } p_{\theta}(\vec{x} | \vec{h}) \right ] \\
&= H(\vec{x} | \vec{h}) + \mathbb{E}_{p(\vec{h})} [D_{\text{KL}}[p(\vec{x} | \vec{h}) || p_{\theta}(\vec{x} | \vec{h})]] \\
&\geq H(\vec{x} | \vec{h}).
\end{align}
This enables us to bound the mutual information as follows:
\begin{align}
I(\vec{x}; \vec{h}) &= H(\vec{x}) - H(\vec{x} | \vec{h}) \\
&\geq H(\vec{x}) - H\left(p(\vec{x} | \vec{h}), p_{\theta}(\vec{x} | \vec{h})\right).
\end{align}
This is similar to the density estimation approach described in Section \ref{subsec:mi_estimation}, however instead of estimating $p(\vec{h}^{\epsilon})$ we are estimating $p(\vec{x} | \vec{h})$. The gap between the true conditional entropy and the conditional cross entropy is given by the KL divergence between the true conditional distribution $p(\vec{x} | \vec{h})$ and our approximating distribution $p_{\theta}(\vec{x} | \vec{h})$ averaged over $\vec{h}$. Therefore the stronger our density model, the better the approximation to the conditional entropy will be. In practice we use an empirical estimate of the conditional cross entropy by averaging across $T$, a held-out test set of $(\vec{x}, \vec{h})$ pairs, :
\begin{align}
H^*(\vec{x} | \vec{h}) &= H^*\left(p(\vec{x} | \vec{h}), p_{\theta}(\vec{x} | \vec{h})\right) \\
 &= -\frac{1}{|T|}\sum\nolimits_{(\vec{x}, \vec{h}) \in T} \text{log } p_{\theta}(\vec{x} | \vec{h})
\end{align}
We could take this a step further, and directly estimate the mutual information by using an unconditional model $p_{\theta}(\vec{x})$ to estimate the marginal data distribution $p(\vec{x})$ and hence the entropy $H(\vec{x})$. However, we don't typically need the absolute value of the mutual information for our analyses, but just the trends in how the mutual information changes between different network layers and settings.  Thus, we directly report the estimated negative cross entropy (NCE), $-H^*(\vec{x}|\vec{h})$, which is equivalent to the mutual information bound, up to the constant $H(\vec{x})$.

\section{Experiments}\label{sec:experiments}\newcommand{\subplotwidth}{0.94}
\begin{figure*}
   \centering
  \resizebox{\textwidth}{!}{%
  \begin{tabular}{ccccc}
    \rotatebox{90}{\hspace{2em}\small MNIST}
    \includegraphics[height=1in]{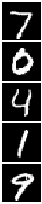}&
    \includegraphics[height=1.184in]{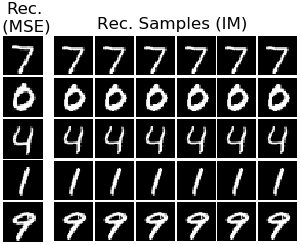}&
    \includegraphics[height=1.184in]{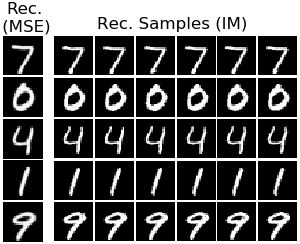}&
    \includegraphics[height=1.184in]{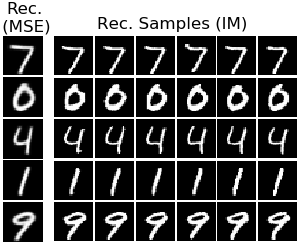}&
    \includegraphics[height=1.184in]{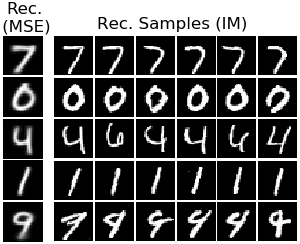}\\
    \rotatebox{90}{\hspace{2em}\small SVHN}
    \includegraphics[height=1in]{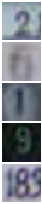}&
    \includegraphics[height=1in]{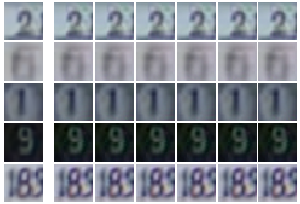}&
    \includegraphics[height=1in]{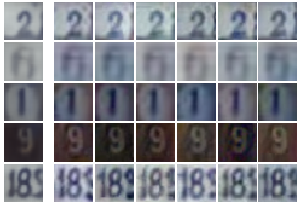}&
    \includegraphics[height=1in]{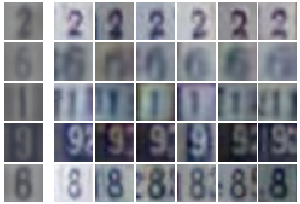}&
    \includegraphics[height=1in]{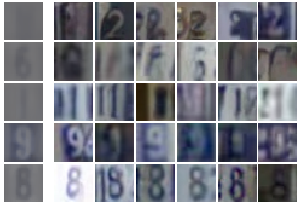}\\
    \rotatebox{90}{\hspace{2em}\small CIFAR}
    \includegraphics[height=1in]{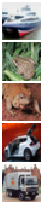}&
    \includegraphics[height=1in]{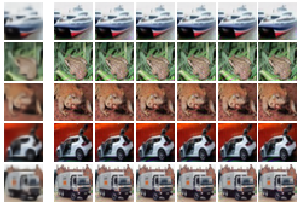}&
    \includegraphics[height=1in]{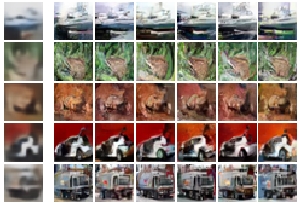}&
    \includegraphics[height=1in]{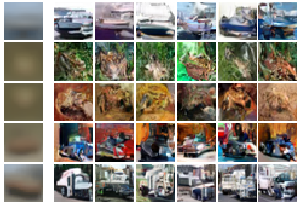}&
    \includegraphics[height=1in]{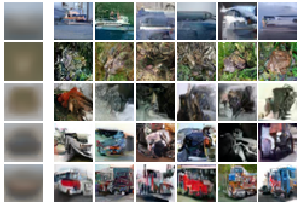}\\
    (a) Input&(b) CONV1&(c) CONV2&(d) FC3&(e) LOGITS\\
  \end{tabular}
  }
  \caption{{\bf Inverting network layers:} We compare both inversion model samples and MSE reconstructions across different layers of the neural network. For MNIST, even the logits retain much of the information about the original inputs. For SVHN and CIFAR the CONV1 and CONV2 layers appear to retain most of the information in the original images, while the FC3 layer becomes invariant to many low and mid-level changes.}
  \label{fig:rec_samples}
\end{figure*}
\subsection{Inverting the layers of image classifiers}\label{sec:inverting_the_layers}
We first explore the use of inversion models to explore the invariances and abstractions learned at each layer in a convolutional neural network.

We trained classifiers on three image datasets: MNIST \citep{lecun1998gradient}, SVHN \citep{netzer2011} and CIFAR-10 \citep{krizhevsky09} achieving test accuracies of 99.6\%, 93.3\% and 81.6\% respectively. Following Huang et al. \citeyearpar{HuangLPHB17} we used ReLU-convolutional networks consisting of two convolutional layers with max pooling, one fully connected layer, and a linear layer that outputs the predicted logits for each class. We refer to these layers as CONV1, CONV2, FC3 and LOGITS respectively. For each dataset and each classifier layer we trained separate PixelCNN++ inversion models. The architectural and training details are described in appendix A.

\boldpar{Samples} Figure \ref{fig:rec_samples} shows samples from trained inversion models along with reconstructions from models trained using the MSE loss from Equation \ref{eq:mse_loss}. For SVHN and CIFAR the sample variability increases significantly from lower to higher layers. This is consistent with the increasing blurriness of the MSE reconstructions and confirms the expectation that input information is increasingly discarded in successive network layers. For MNIST the reconstructions are visually similar right up to the output layer, indicating that strong invariance in intermediate layers is not a requirement for good performance on this dataset. 


\begin{figure*}[t]
	\centering
	\begin{subfigure}[b]{0.1\textwidth}
		\centering
		\includegraphics[height=1.175in]{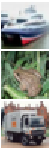}
		\caption{Inputs}
	\end{subfigure}%
	\begin{subfigure}[b]{0.45\textwidth}
		\centering
		\includegraphics[height=1.175in]{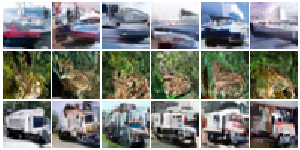}
		\caption{Top-k inversion model samples}
	\end{subfigure}%
	\begin{subfigure}[b]{0.45\textwidth}
		\centering
		\includegraphics[height=1.175in]{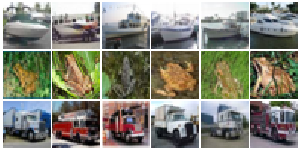}
		\caption{Nearest neighbor training examples}
	\end{subfigure}%
	\caption{{\bf Comparing inversion model samples to nearest neighbors:}
		For each input image we sample 1024 images using an FC3 inversion model, and show the top-k by $L_1$ distance.  We compare these to the top-k nearest neighbors from the training set calculated in the same way.}\label{fig:nearest_neighbours}
	\begin{subfigure}[b]{0.33\textwidth}
		\centering
		\includegraphics[width=\subplotwidth\linewidth]{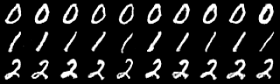}
		\caption{MNIST}
	\end{subfigure}%
	\begin{subfigure}[b]{0.33\textwidth}
		\centering
		\includegraphics[width=\subplotwidth\linewidth]{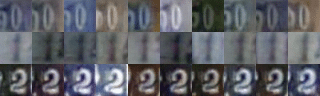}
		\caption{SVHN}
	\end{subfigure}%
	\begin{subfigure}[b]{0.33\textwidth}
		\centering
		\includegraphics[width=\subplotwidth\linewidth]{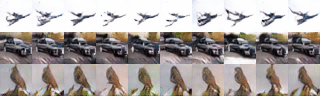}
		\caption{CIFAR}
	\end{subfigure}%
	\caption{{\bf SGAN samples conditioned on FC3:} The SGAN samples are less diverse than the inversion model samples and the nearest neighbors shown in Figure \ref{fig:nearest_neighbours}.  In the SGAN model, this diversity is controlled by an entropy loss term with a tunable weight, making it difficult to calibrate it in a principled way.}
	\label{fig:sgan_samples}
\end{figure*}
\begin{table*}[t]
	\centering
	\footnotesize
	\def\arraystretch{1.1}%
	\begin{tabular}{@{}lccccccccc@{}}
		\toprule
		& \multicolumn{3}{c}{\textbf{MNIST}} & \multicolumn{3}{c}{\textbf{SVHN}} & \multicolumn{3}{c}{\textbf{CIFAR}} \\ \cmidrule(lr){2-4}\cmidrule(lr){5-7}\cmidrule(lr){8-10}
		& CONV1      & CONV2     & FC3       & CONV1     & CONV2     & FC3       & CONV1      & CONV2     & FC3       \\ \midrule
		1NN       & 1.25e-2    & 4.95e-2   & 1.22e-1   & 5.35e-2   & 6.74e-2   & 2.92e-1   & 4.11e-2    & 9.56e-2   & \textbf{5.28e-1}   \\
		IM-S & 7.68e-4    & 1.65e-2   & 1.35e-1   & 1.02e-3   & 2.80e-2   & 3.22e-1   & 4.02e-3    & 6.43e-2   & 7.39e-1   \\
		IM-NN     &  \textbf{6.29e-4}          & \textbf{1.40e-2}          &  \textbf{9.52e-2}         &  \textbf{9.40e-4}         &    \textbf{2.43e-2}         &  \textbf{2.50e-1}          &    \textbf{3.47e-3}        &    \textbf{5.73e-2}        &    5.79e-1       \\ \bottomrule
	\end{tabular}
	\caption{{\bf Comparing images in representation space:}  For 500 test samples from each dataset we compute the average $L_1$ distance at various network layers between the input image and: (1NN) the nearest neighbor training set example, (IM-S) a random inversion sample or (IM-NN) the closest of 10 inversion model samples.  We can see that the IM-NN distance is the closest for all datasets and layers except the FC3 layer of CIFAR10.}\label{tab:h_distance}
\end{table*}

For all datasets, CONV1 samples are almost indistinguishable from the inputs. CONV2 reconstructions also preserve information about the locations, styles and colors of objects and digits, but are more variable with respect to finer details. There is a distinct increase in reconstruction variability from CONV2 to FC3, particularly on the CIFAR dataset. However color information is preserved in FC3, along with object structures and scene textures. A surprising amount of information is retained even in the networks' logit predictions; this is particularly evident in the MNIST reconstructions for which style and orientation information is preserved. This is consistent with the dark knowledge hypothesis described by Hinton et al. \citeyearpar{HintonVD15} that suggests that the particular output probabilities that a model assigns to its inputs provides a rich characterization of the similarity between examples.

\begin{figure*}[t]
	\centering
	\begin{subfigure}[b]{0.08\textwidth}
		\centering
		\includegraphics[height=0.895in]{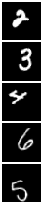}
		\caption{Input}
		\label{fig:MNIST_mutual_info}
	\end{subfigure}%
	\begin{subfigure}[b]{0.31\textwidth}
		\centering
		\includegraphics[height=1.05in]{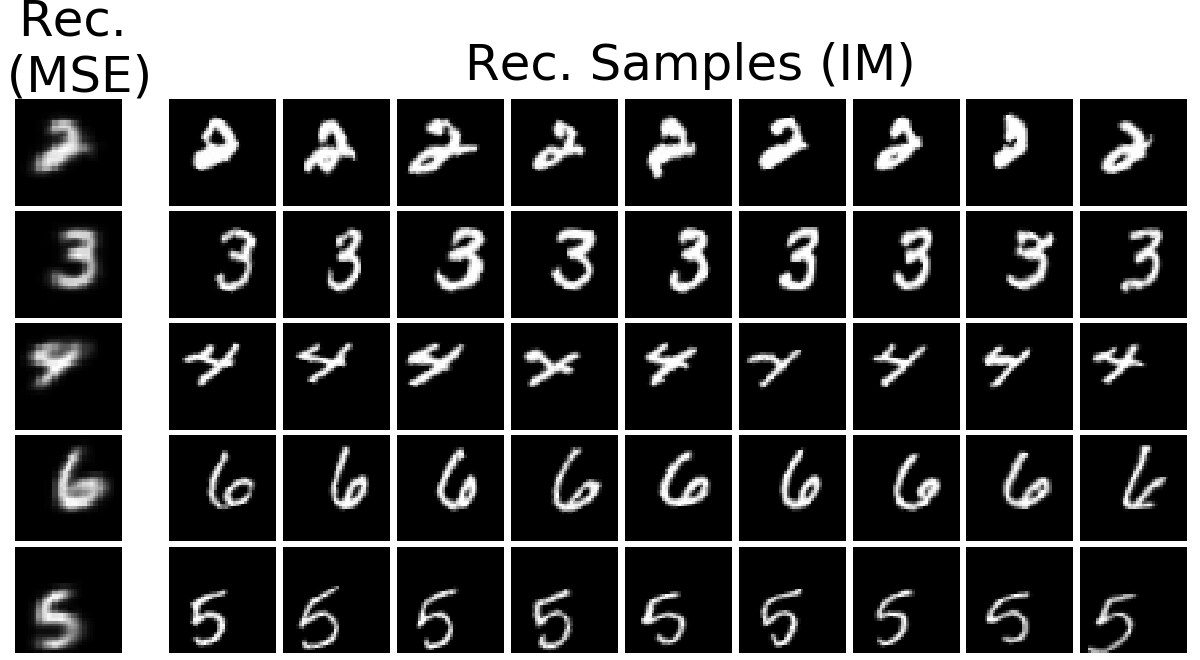}
		\caption{Fully connected}
		\label{fig:MNIST_mutual_info}
	\end{subfigure}%
	\begin{subfigure}[b]{0.31\textwidth}
		\centering
		\includegraphics[height=1.05in]{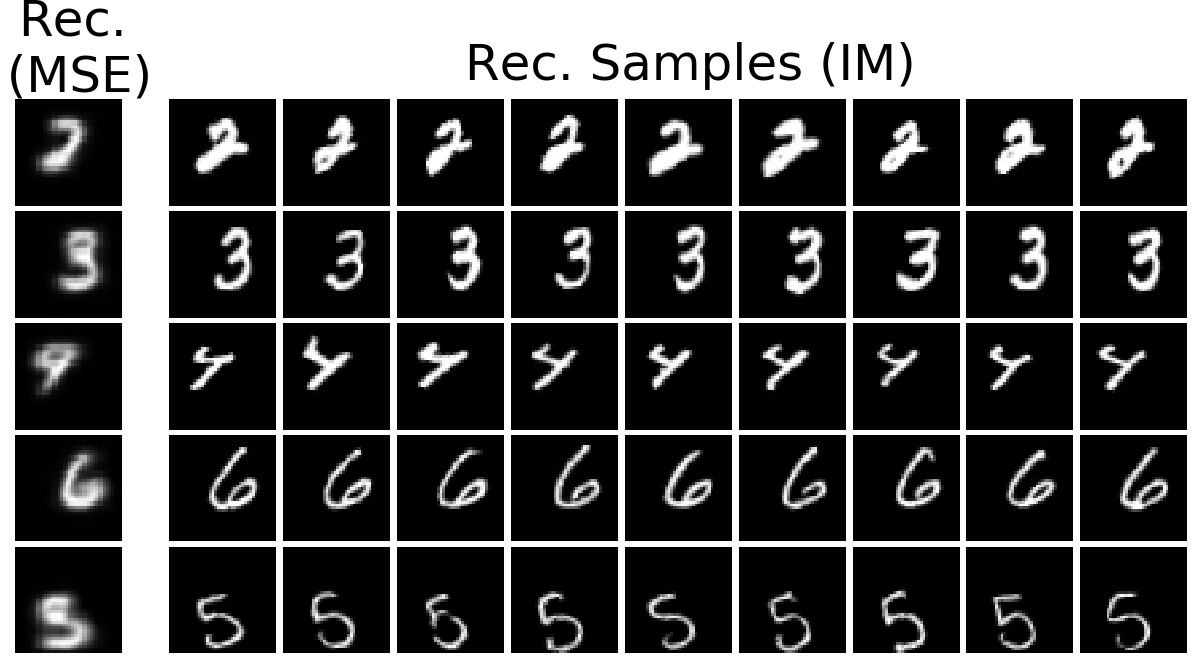}
		\caption{Conv (no Global Pooling)}
		\label{fig:SVHN_mutual_info}
	\end{subfigure}%
	\begin{subfigure}[b]{0.31\textwidth}
		\centering
		\includegraphics[height=1.05in]{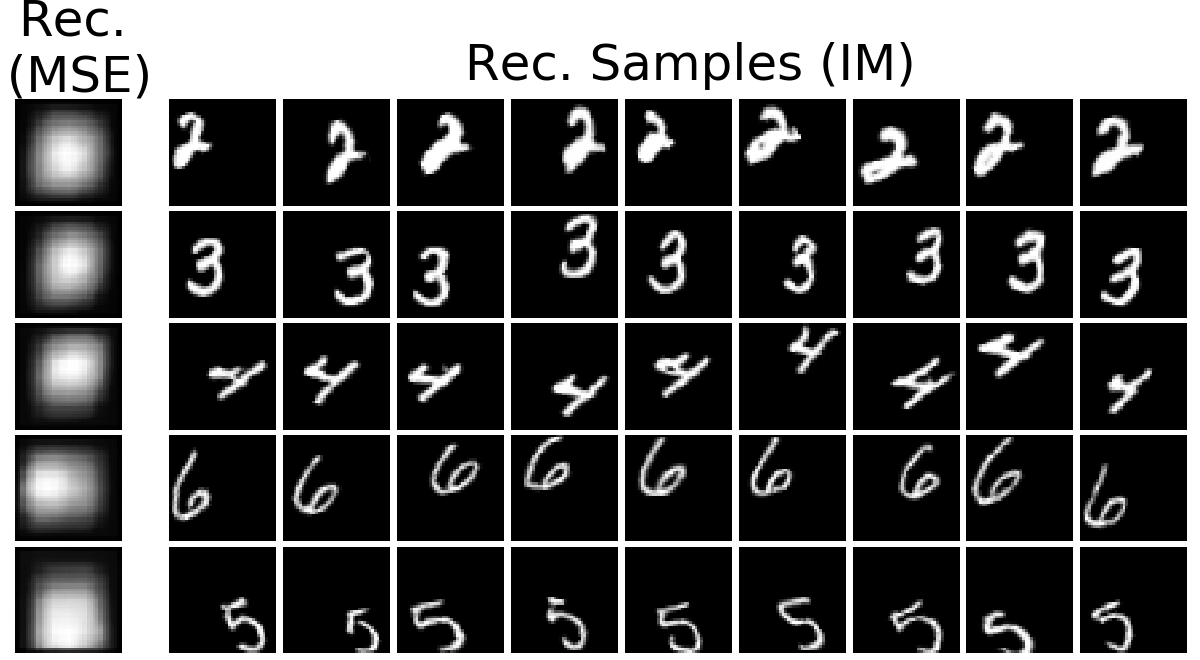}
		\caption{Conv (with Global Pooling)}
		\label{fig:SVHN_mutual_info}
	\end{subfigure}%
	\caption{{\bf Architecture comparison:}  Inversion model samples and MSE reconstructions from the FC3 layers of three network architectures trained on affNIST. The network with global pooling learns translational invariance, while the others do not. Furthermore, the inversion samples indicate that the style and orientation of the digits are retained in the global pooling network, an observation that cannot be gleaned from the MSE reconstructions.}\label{fig:affNIST_recs}
\end{figure*}

%
%

\boldpar{Mutual information} Figure \ref{fig:classifier_MI}a shows lower bounds on the negative cross entropy at the different layers in the neural network, as discussed in section~\ref{sec:bounding_the_mutual_information}. In general, the mutual information bounds decrease through successive layers of the network, indicating a general loss of information about the input. The biggest reduction in mutual information by far is between CONV1 and CONV2, which is surprising as the biggest jump in variability for samples from the inversion models appears to occur in later layers. It is possible that our intuitions about visual information are poorly calibrated, and that we underestimate the amount of information present in high-frequency pixel detail. This is supported by the popularity of perceptual loss metrics in generative vision applications \citep{JohnsonAF16, DosovitskiyB16}, that aim to characterize the differences between images in a feature space better aligned with human perception.

\boldpar{Comparison to nearest neighbors and SGAN} In order to evaluate how well the inversion models capture the distribution $p(\vec{x} | \vec{h})$ we pass the generated inversion samples, $\vec{\hat{x}}$, back through the image classifier to generate the hidden states $\vec{\hat{h}}$ for these generated images. Table~\ref{tab:h_distance} shows the results of doing this for 500 test set images and computing the average $L_1$ distance between $\vec{h}$ and $\vec{\hat{h}}$.  We also perform the same calculation using the nearest neighbor images in the training set, instead of generated samples.  We report the distances for a single random sample from an inversion model (IM-S) as well as for the closest of 10 random samples (IM-NN). We find that for a single random sample, the FC3 $L_1$ distance is smaller than the that of the nearest training set example for CONV1 and CONV2, but not for FC3. However the IM-NN distances is smaller for all layers on the MNIST and SVHN datasets, but not on the CIFAR dataset. Figure \ref{fig:nearest_neighbours} shows the 6 nearest training examples, and the 6 closest inversion model samples from a collection of 1024, for a selection of input images. 

For comparison, in Figure~\ref{fig:sgan_samples} we also show samples from an SGAN conditioned on the FC3 layer of an equivalent CNN (reproduced from~\citep{HuangLPHB17}).  We can see that the variability of these samples is much lower than both the samples from our model and from the nearest neighbors in Figure~\ref{fig:nearest_neighbours}c.  In fact, in order to prevent the SGAN from collapsing to a deterministic function, the authors added an explicit entropy term to the loss function.  As the weight on this loss term is a tunable hyperparameter, there's no principled way to set this to ensure that the variability is well calibrated.  In contrast, since we are using an autoregressive model which is trained directly to maximize the likelihood of the data, we need no such tunable hyperparameter, and can expect the variance to be better calibrated.  It's also worth noting the our use of the autoregressive model enables estimation of the mutual information, as discussed above, a capability not provided by the SGAN model.

\subsection{Comparing network architectures using inversion models}\label{sec:case_study}

Another practical application of inversion models is to facilitate analysis of network architectures by revealing the invariances present in various network layers. If we know that our networks are not learning the desired invariances, we can take steps to modify our architecture. As a case study, we analyze a design choice for convolutional architectures that has become increasingly popular: global spatial pooling. Global pooling layers aggregate information from all spatial locations, greatly reducing the number of parameters in network architectures. They have been found to help reduce overfitting, and have largely replaced fully connected layers in the final processing steps in modern image architectures \citep{LinCY13, HeZRS16}. As an additional point of comparison we include a network that replaces convolutional layers with fully connected layers. 

We train supervised models on the affNIST dataset\footnote{Available at \url{https://www.cs.toronto.edu/~tijmen/affNIST/}}. affNIST consists of MNIST digits with random affine transformations on a $40 \times 40$ canvas. These transformations increase the need for invariance in network representations in comparison to standard MNIST. We trained three supervised networks: the first is identical to that used on MNIST in Section \ref{sec:inverting_the_layers}, the second applies global max pooling to the CONV2 feature maps and passes the resulting vector to FC3. The final network replaces convolutional layers with fully connected layers with 2048 units. The network with global pooling performs best, achieving 98.9\% accuracy compared to 98.7\% for the version without global pooling and 95.0\% for the fully connected network. We trained PixelCNN++ inversion models to invert FC3 representations for the supervised networks, using the same architecture as for the MNIST model in section \ref{sec:inverting_the_layers}.


For the fully connected network we obtain a relative mutual information lower bound of $-715.17$ nats. For the convolutional networks with and without global pooling we obtain $-704.01$ and $-699.55$ nats respectively. This indicates that more input information is preserved in the layers of the convolutional networks than the fully-connected network, and that the global pooling layer discards an estimated $\sim 4.5$ nats of information about the inputs.

Figure \ref{fig:affNIST_recs} shows samples from the inversion models along with MSE reconstructions. The samples indicate that for the network with global pooling the translation of the digit is not preserved in FC3, whereas for the network without global pooling and the fully connected network the digit's location is preserved. These results are reinforced by the MSE reconstructions, which are very blurry for the network with global-pooling. It should be noted that it is not possible to tell from the MSE reconstructions that the global-pooling network preserves style and rotation information about the digit, whereas samples from the inversion model indicate that this is the case. It is surprising that the models without global pooling do not achieve translation invariance by FC3, and it may explain their relatively worse performance, as the networks must learn this invariance in the final linear layer.

\subsection{Training dynamics}\label{sec:compression_dynamics}
\begin{figure*}
	\centering
    \begin{subfigure}[b]{0.3\textwidth}
	    \centering
        \includegraphics[height=1.5in]{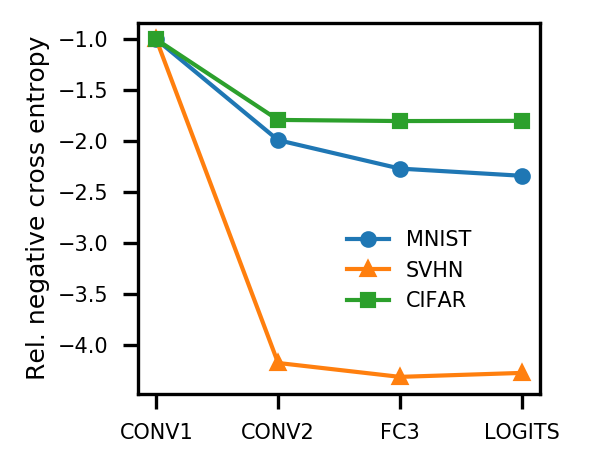}
        \caption{All Datasets}
        \label{fig:MNIST_mutual_info}
    \end{subfigure}%
    \begin{subfigure}[b]{0.7\textwidth}
    	\centering
        \includegraphics[height=1.5in]{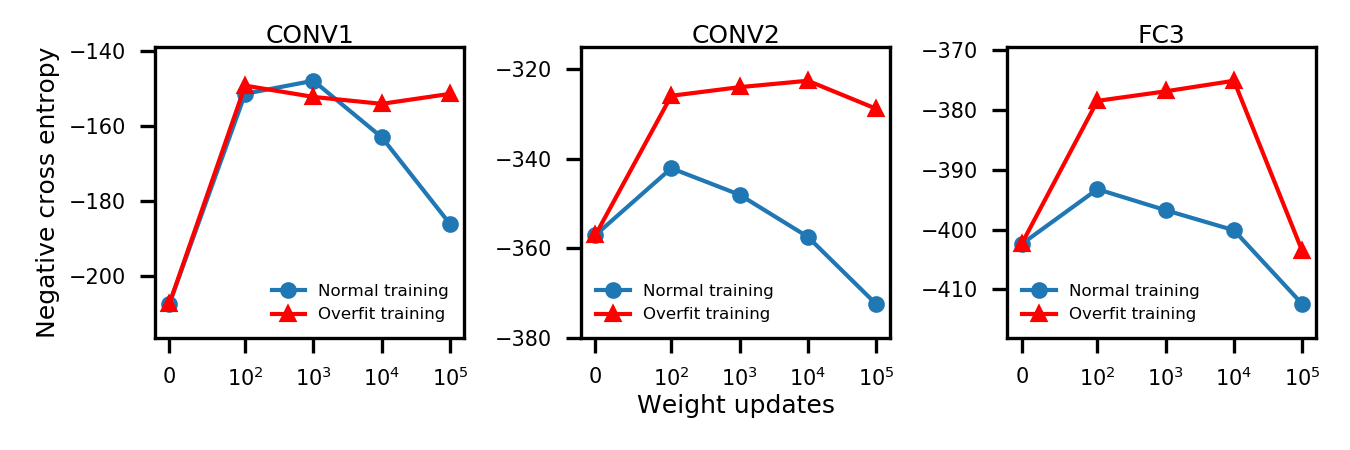}
        \caption{MNIST across training steps}
        \label{fig:SVHN_mutual_info}
    \end{subfigure}%
    \caption{\textbf{Mutual information analysis:} (a) Negative Cross Entropy (NCE) across datasets and layers, expressed relative to the magnitude of the CONV1 cross entropy. (b) Absolute NCE in nats for network layers over the course of training for regular (blue) and intentionally overfitted (red) classifiers on MNIST. The initial increase and subsequent decrease in NCE over the course of training for the regular model is consistent with the expansion and compression phases predicted in previous work \citep{Shwartz-ZivT17}. The NCE is considerably higher for the overfitted network.}\label{fig:classifier_MI}
\end{figure*}

Inversion models can also be used to better understand the compression dynamics of neural network training.  These dynamics were recently studied by Schwartz-Ziv and Tishby \citeyearpar{Shwartz-ZivT17} who examined the mutual information between inputs and intermediate layers over the course of SGD optimization. Their findings suggested that there exist two distinct phases of training: an expansion phase in which networks increase the mutual information between inputs and hidden layers and a compression phase in which the mutual information is reduced as information is filtered out. More recently Saxe et al. \citeyearpar{michael2018} provided evidence that the compression phase only occurs when saturating non-linearities such as tanh's are used, and that no compression is present for ReLU networks. In both cases, the mutual information was estimated using either discretization or non-parametric methods, each of which have issues in terms of scalability to high dimensionality features as discussed in Section \ref{subsec:mi_estimation}. Instead, we can use inversion models to examine these claims for the larger networks that we employ.

Using the MNIST classifier described in Section \ref{sec:inverting_the_layers} we trained inversion models on representations established after $0$, $10^2$, $10^3$, $10^4$ and $10^5$ weight updates. We use this relatively coarse view, rather than the finer view taken by Schwartz-Ziv and Tishby due to the computational expense of training inversion models.  For the supervised MNIST network, which takes about $5 \times 10^4$ weight updates to converge, this range is fairly representative of the training process. In order to investigate the connections between mutual information and generalization we additionally trained an overfitted network by using only 100 training examples and removing dropout from the the classifier. We use equivalent inversion models to the ones described in section \ref{sec:inverting_the_layers}.

Figure \ref{fig:classifier_MI}b shows the negative cross entropy of inversion models trained at different network layers over the course of training. As discussed in Section \ref{sec:bounding_the_mutual_information} this is equal to a lower bound on the mutual information up to a constant. The results for normal training are shown in blue, and the results for the overfitted network are shown in red. Our main observation is that in the normal training regime for all layers of the network the lower bound on the mutual information initially increases, and the decreases significantly over the course of training. We therefore see a reproduction of the main findings of Schwartz-Ziv and Tishby for ReLU-convolutional networks. This apparent contradiction of the findings of Saxe et al. \citeyearpar{michael2018} can potentially be explained by their use of non-parametric methods to estimate the mutual information. Our results additionally indicate that the mutual information is considerably higher for the overfitted network than for the well-regularized network, which supports the notion that compression in network layers has an important role in a model's generalization performance.

\section{Discussion}\label{sec:discussion}In this work we present a method for the inversion of supervised representations that uses flexible autoregressive neural density models to express a distribution over inputs given an intermediate representation. Our method has two benefits: it facilitates visualisation of model invariances, thus enabling analysis of architectural choices. Secondly it provides a scalable quantitative estimate of the amount of information preserved by a network. One difficulty is that density estimation is challenging in higher dimensions, and that we don't know how well a given model represents the true density. However, as neural density models improve, so too does our method, and we will be able to achieve tighter mutual information bounds and more representative samples.

There are number of directions for future work, including the comparison of representations learned
using different optimizers, or in different training regimes. Additionally it would be of interest to
analyze the effect of training techniques such as dropout or batch normalization, or of architectural
choices such as residual connections on representation and compression.

\subsubsection*{Acknowledgements}

The work of Christopher Williams is supported by EPSRC grant EP/N510129/1 to the Alan Turing Institute. Charlie Nash is supported by the Centre for Doctoral Training in Data Science, funded by EPSRC grant EP/L016427/1 and by a Microsoft Azure Research Award.

\bibliography{bibliography}

\newpage
\section*{Appendix A: Experimental details}\label{App:A}
\boldpar{Image classifiers} For each image classifier in Section \ref{sec:inverting_the_layers} we used the same network structure: Two layers of convolution-ReLU-max-pooling, each with kernels of size 5, and no padding, followed by one fully connected layer, and a final linear layer that outputs the unnormalized softmax probabilities. For MNIST we used 32 filters in each convolutional layer, and for SVHN and CIFAR we used 64 and 128 filters for CONV1 and CONV2 respectively. The fully connected layer takes the vectorized feature maps of CONV2 and maps to FC3 which has 256 units for all datasets. We used dropout at every layer, and Adam optimizer with learning rate $3 \times 10^{-4}$. We trained the networks for a maximum of 250000 steps, and used early stopping with respect to the validation accuracy.

For the AffNIST experiments in Section \ref{sec:case_study} we use three image classifiers, one with global spatial pooling, one without global spatial pooling, and a network that replaces convolutional layers with fully connected layers. The network without global spatial pooling is identical to the MNIST network used in Section \ref{sec:inverting_the_layers}. The network with global spatial pooling is identical except that it applies global max pooling to the CONV2 feature maps. The output of this operation is a vector with the same number of dimensions as there are channels in the CONV2 feature maps. This vector is then connected using a fully-connected layer to the 256 unit FC3 features. The fully connected network replaces CONV1 and CONV2 with fully connected layers each with 2048 units. The training details are the same as for the other image classifiers.

\boldpar{PixelCNN++} For all inversion networks we used the PixelCNN++ architecture detailed in \citep{SalimansKCK17}. The architecture consists of six blocks of residual layers, with spatial downsampling using strided convolutions between the first, second and third blocks, and spatial upsampling using strided transpose convolutions between the fourth, fifth and sixth blocks. In order to preserve high resolution information skip connections are employed between corresponding downsampling and upsampling blocks. In order to reduce the cost of training the models, we used three residual layers in each block rather than the five specified in the original architecture. We use 64 filters in the convolutional layers for MNIST and 196 for SVHN and CIFAR. For SVHN and CIFAR we used the discretized mixture of logistics described in \citep{SalimansKCK17} with 10 mixture components for the conditional pixel likelihood. For MNIST we used a 256-way softmax distribution over the discrete pixel values as in the original PixelCNN \citep{OordKK16} as we found it to be much more effective in practice. 

We used weight normalization with data-dependent initialization \citep{SalimansK16}, and trained our models using Adam optimizer with initial learning rate $10^{-3}$ and a learning rate decay of 0.9999 for a maximum of 250000 weight updates. Again we used early stopping, but found that in general the validation performance continued to improve for the duration of training. To condition on vector representations FC3 and LOGITS we linearly projected the context vector to biases that are added to feature maps in each residual layer. For spatial representations CONV1 and CONV2 we resize the context to match the PixelCNN++ feature maps and use $1 \times 1$ convolutions to project to spatially-structured biases that are added to the feature maps. We used a single dropout layer in each PixelCNN++ residual block for all networks. For CONV1 inversion models on SVHN and CIFAR we used dropout rate 0.1, and for all other networks we used dropout rate 0.5. We applied an additional dropout layer with dropout rate 0.2 to the outputs of the linear projection of the context for the CIFAR-FC3 inversion model.

\section*{Appendix B: More details on mutual information estimation in neural networks}\label{app:B}
Here we describe some alternative approaches to mutual information estimation in neural networks not covered in Section \ref{subsec:mi_estimation}.

\boldpar{Discretization} Schwartz-Ziv and Tishby \citeyearpar{Shwartz-ZivT17} discretize tanh activations in a fully-connected neural network each into 30 equally sized bins to form a discrete empirical distribution $p(\vec{h} | \vec{x})$. In experiments with a known distribution of discrete inputs $p(\vec{x})$, they exactly compute the mutual information between the inputs and the discretized layer activations by averaging over settings of $\vec{x}$ and $\vec{h}$. 
\begin{align}
I(\vec{x}; \vec{h}) = \sum\nolimits_{\vec{x}, \vec{h}} p(\vec{x}, \vec{h}) \ \text{log}\left(\frac{p(\vec{x}, \vec{h})}{p(\vec{x})p(\vec{h})}\right)
\end{align}
Saxe et al., \citeyearpar{michael2018} note that the networks of interest do not operate on the discretized values, and that the binning is used solely for mutual information calculations. Moreover, there are many possible ways of binning potentially unbounded activations such as ReLUs, and the choice can significantly impact the mutual information estimates.

\boldpar{Non-parametric entropy estimation} Saxe et al., \citeyearpar{michael2018} similarly obtain an approximate bound on the mutual information by estimating the entropy of activations with additive noise $H(\vec{h}^\epsilon)$. They use the estimator of Kraskov et al. \citeyearpar{kraskov2004} that makes use of distances between nearest neighbours in a collection of samples. 
The entropy estimator is:
\begin{align}
\hat{H}(\vec{h}) &= \frac{D}{N}\sum\nolimits_i \text{log}(r_i + \epsilon) + \frac{D}{2}\text{log}(\pi) \\
&- \text{log}\Gamma(D/2 + 1) + \psi(N) - \psi(k),
\end{align}
where $D$ is the dimensionality of $\vec{h}$, $N$ is the number of samples, $r_i$ is the distance between sample $i$ and its $k$'th nearest neighbour, $\Gamma$ is the Gamma function, and $\psi$ is the digamma function. 
As with the KDE-based approach described in Section \ref{subsec:mi_estimation}, this non-parametric estimate may be problematic for analysis of network layers with very many units. 
\end{document}